\documentclass{article}

\usepackage[preprint]{nips_2018}

\usepackage[utf8]{inputenc} 
\usepackage[T1]{fontenc}    
\usepackage{hyperref}       
\usepackage{url}            
\usepackage{booktabs}       
\usepackage{amsfonts}       
\usepackage{nicefrac}       
\usepackage{microtype}      
\usepackage{graphicx}
\usepackage{listings}

\newcommand{\code}[1]{\texttt{#1}}

\title{Mixed-Precision Training for NLP and Speech Recognition with OpenSeq2Seq}

\author{
  Oleksii Kuchaiev, Boris Ginsburg, Igor Gitman\thanks{work done while the author was an intern at NVIDIA}, \\
  \textbf{Vitaly Lavrukhin, Jason Li, Huyen Nguyen, Carl Case, Paulius Micikevicius} \\
  NVIDIA \\
  Santa Clara, CA 95051\\
  \texttt{\{okuchaiev,bginsburg,igitman,}\\\texttt{vlavrukhin,jasoli,chipn,carlc,pauliusm\}@nvidia.com} \\
}

\begin{document}

\maketitle

\begin{abstract}
  We present OpenSeq2Seq -- a TensorFlow-based toolkit for training sequence-to-sequence models that features distributed and mixed-precision training. Benchmarks on machine translation and speech recognition tasks show that models built using OpenSeq2Seq give state-of-the-art performance at 1.5-3x less training time. OpenSeq2Seq currently provides building blocks for models that solve a wide range of tasks including neural machine translation, automatic speech recognition, and speech synthesis.
\end{abstract}

\section{Introduction}
\label{intro}
The sequence-to-sequence (seq2seq) paradigm \citep{cho2014learning} has been successfully used for tasks that traditionally require a sequential encoder and a sequential decoder such as machine translation \citep{wu2016google}, abstractive summarization \citep{rush2015neural}, and automatic speech recognition (ASR) \citep{chan2015listen, battenberg2017exploring}. However, seq2seq models can be used for other tasks as well. For example, a neural network to solve a sentiment analysis task might consist of an RNN encoder and a softmax linear decoder. An image classification task might need an convolutional encoder and a softmax linear decoder. A model that translates from English to multiple languages might have one encoder with multiple decoders.

There have been a number of toolkits that use the seq2seq paradigm to construct and train models to solve various tasks. Some of the most popular include Tensor2Tensor \citep{tensor2tensor}, seq2seq \citep{Britz:2017}, OpenNMT \citep{opennmt}, and fairseq \citep{gehring2017convs2s}. The first two are based on TensorFlow \citep{abadi2016tensorflow} while the last two are based on PyTorch \citep{paszke2017automatic}. These frameworks feature a modular design with many off-the-shelf modules that can be assembled into desirable models, lower the entrance barrier for people who want to use seq2seq models to solve their problems, and have helped push progress in both AI research and production. 

OpenSeq2Seq builds upon the strengths of these existing frameworks with additional features to reduce the training time and make the API even easier to use. We chose to work with TensorFlow. We created OpenSeq2Seq with the following goals in mind:
\begin{itemize}
    \item Modular architecture to allow easy assembling of new models from available components.
    \item Support for mixed-precision training \citep{micikevicius2017mixed} , that utilizes Tensor Cores introduced in NVIDIA Volta GPUs.
    \item Fast, simple-to-use, Horovod-based distributed training via data parallelism, supporting both multi-GPU and multi-node.
\end{itemize}
OpenSeq2Seq is open-source at: \url{https://github.com/NVIDIA/OpenSeq2Seq}.

\section{Modular architecture}
\label{modular}
OpenSeq2Seq was designed for extensibility and modularity. It provides core abstract classes from which users can inherit their own modules: \code{DataLayer},  \code{Model}, \code{Encoder}, \code{Decoder} and \code{Loss}.

At a high level, the \code{Encoder} consumes data processed by \code{DataLayer} and produces a representation; while the \code{Decoder} consumes that representation and produces data and/or output. We assume that any encoder can be combined with any decoder, thus improving flexibility and simplicity of experimentation. It is possible to have a model consisting of only an encoder, or having more than one encoder and/or decoder.

An OpenSeq2Seq model is described by a Python configuration file that specifies parts of the model (i.e. data layer, encoder, decoder, and loss function) and their configuration hyperparameters (i.e. regularization, learning rate, dropout). A configuration file to create an GNMT \citep{wu2016google} model for machine translation might look like this:
\begin{lstlisting}[basicstyle=\small]
base_params = {
  "batch_size_per_gpu": 32,
  "optimizer": "Adam",
  "lr_policy": exp_decay,
  "lr_policy_params": {
    "learning_rate": 0.0008,
},
"encoder": GNMTLikeEncoderWithEmbedding,
  "encoder_params": {
    "core_cell": tf.nn.rnn_cell.LSTMCell,
    ...
    "encoder_layers": 7,
    "src_emb_size": 1024,
  },
"decoder": RNNDecoderWithAttention,
  "decoder_params": {
    "core_cell": tf.nn.rnn_cell.LSTMCell,
   ...
  },
"loss": BasicSequenceLoss,
   ...
}
\end{lstlisting}
To create a model, OpenSeq2Seq provides \code{run.py} script which takes as arguments the model's configuration file and the execution mode (\textit{train}, \textit{eval}, \textit{train\_eval} or \textit{infer}).

Currently, OpenSeq2Seq provides configuration files to create models for machine translation (GNMT, ConvS2S, Transformer), speech recognition (Deep Speech 2, Wav2Letter), speech synthesis (Tacotron 2), image classification (ResNets, AlexNet), language modeling (LSTM-based), and transfer learning for sentiment analysis. These are stored in the folder \code{example\_configs}. You can create a new model configuration using the modules available in the toolkit with basic knowledge in TensorFlow. It’s also straightforward to write a new module or to modify an existing module. 

OpenSeq2Seq provides a variety of data layers that can process popular datasets, including WMT for machine translation, WikiText-103 \citep{merity2016pointer} for language modeling, LibriSpeech \citep{panayotov2015librispeech} for speech recognition, Stanford Sentiment Treebank \citep{socher2013recursive} and IMDB \citep{maas-EtAl:2011:ACL-HLT2011} for sentiment analysis, LJ Speech dataset \citep{ito2017lj} for speech synthesis, and more.

\section{Mixed-precision training}
\subsection{Mixed-precision algorithm}
Tensor Cores, available on Volta and Turing GPUs, allow matrix-matrix multiplication, the operations at the core of neural network training and inferencing, to be done in both single-precision floating point (FP32) and half-precision floating point (FP16). For training, Tensor Cores provide up to 12x higher peak TFLOPS compared to standard FP32 operations on P100. For inference, that number is 6x \citep{nvidia2017v100}. 

Taking advantage of Tensor Cores' computational power requires models to be trained using FP16 arithmetic. OpenSeq2Seq provides a simple interface to do so via mixed-precision training. When mixed-precision training is enabled, the math is done in FP16, but the results are accumulated and stored in FP32. In current generation GPUs, reduced precision math increases the computational throughput. Mixed-precision also reduces the amount of memory required, allowing users to increase the size of batches or models, which, in turn, increases the learning capacity of the model and reduces the training time. 

To prevent accuracy loss due to the reduced precision, we use two techniques suggested by \cite{micikevicius2017mixed}:
\begin{enumerate}
    \item Automatically scale loss to prevent gradients from underflow and overflow during backpropagation. The optimizer inspects gradients at each iteration and scales the loss for the next iteration to ensure that the values stay within the FP16 range.
    \item Maintain a FP32 copy of weights to accumulate the gradients after each optimizer step.
\end{enumerate}
While having two copies of weights increases the memory consumption, the total memory requirement is often \textit{decreased} because activations, activation gradients, and other intermediate tensors are kept in FP16. This is especially beneficial for models with a high degree of parameter sharing, such as recurrent neural networks.

To enable mixed-precision training in OpenSeq2Seq, simply change dtype parameter of \code{model\_params} to “mixed” in your configuration file. You can enable loss scaling either statically by setting \code{loss\_scale} parameter to the desired number, or dynamically by setting \code{loss\_scaling} parameter to “Backoff” or “LogMax”. You may need to pay attention to the types of the inputs and outputs to avoid mismatched types for certain types of computations. There’s no need to modify the architecture or hyperparameters.

\begin{lstlisting}[basicstyle=\small]
base_params = {
  ...
  "dtype": "mixed",
  # "loss_scale": 10.0, # static loss scaling
  # "loss_scaling": "Backoff", # dynamic loss scaling
}
\end{lstlisting}

\subsection{Mixed-precision optimizer}
\begin{figure*}[t]
\centering
\includegraphics[width=0.7\textwidth]{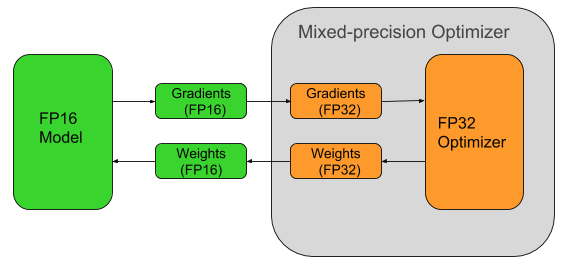}
\caption{Mixed-precision wrapper around TensorFlow optimizers}
\label{fig:MixedPrecisonOptimizer}
\end{figure*}

Our implementation is different from the approach explained in \cite{NVMixed}: instead of using a custom variable getter, we introduce a wrapper around the standard TensorFlow optimizers. The model is created with FP16 -- all variables and gradients are in FP16 by default, except for the layers which are explicitly redefined as FP32 such as data layers or operations on CPU. The wrapper then automatically converts FP16 gradients to FP32 and submits them to TensorFlow optimizer, which updates the master copy of weights in FP32. Updated FP32 weights are converted back to FP16, which are then used by the model in the next forward-backward iteration. Figure \ref{fig:MixedPrecisonOptimizer} illustrates the \code{MixedPrecisionOptimizerWrapper} architecture.

\subsection{Mixed-precision regularization}
Mixed-precision training may require special care for regularization. Consider, for example, weight decay regularization. Given that the weights are commonly initialized with small values, multiplying them with weight decay coefficient - which is usually on the order of $\left[10^{-5}, 10^{-3}\right]$ -- can result in numerical underflow. 

To overcome this problem, the regularizer function is wrapped with \code{mp\_regularizer\_wrapper} function that disables the underlying regularization function for FP16 copy and adds the regularized variables to a TensorFlow collection. This collection will later be retrieved by \code{MixedPrecisionOptimizerWrapper}. The corresponding regularizer functions will be applied to the FP32 copy of the weights to ensure that their gradients always stay in full precision. Since this regularization is not in the loss computation graph, we explicitly call \code{tf.gradients} and add the result to the gradients passed in the \code{compute\_gradients} in the optimizer.
\section{Distributed training with Horovod}
OpenSeq2Seq takes advantage of the two main approaches for distributed training:
\begin{itemize}
    \item Parameter server-based approach (used in native TensorFlow towers)
    \begin{itemize}
        \item Builds a separate graph for each GPU
        \item Sometimes faster for 2 to 4 GPUs
    \end{itemize}
    \item MPI-based approach (used in Uber’s Horovod \citep{sergeev2018horovod})
    \begin{itemize}
        \item Uses MPI and NVIDIA’s NCCL library to utilize NVLINK between GPUs
        \item Significantly faster for 8 to 16 GPUs
        \item Fast multi-node training
    \end{itemize}
\end{itemize}
To use the first approach, you just need to update the configuration parameter \code{num\_gpus} to the number of GPUs you want to use. 

\begin{figure*}[t]
\centering
\includegraphics[width=0.8\textwidth]{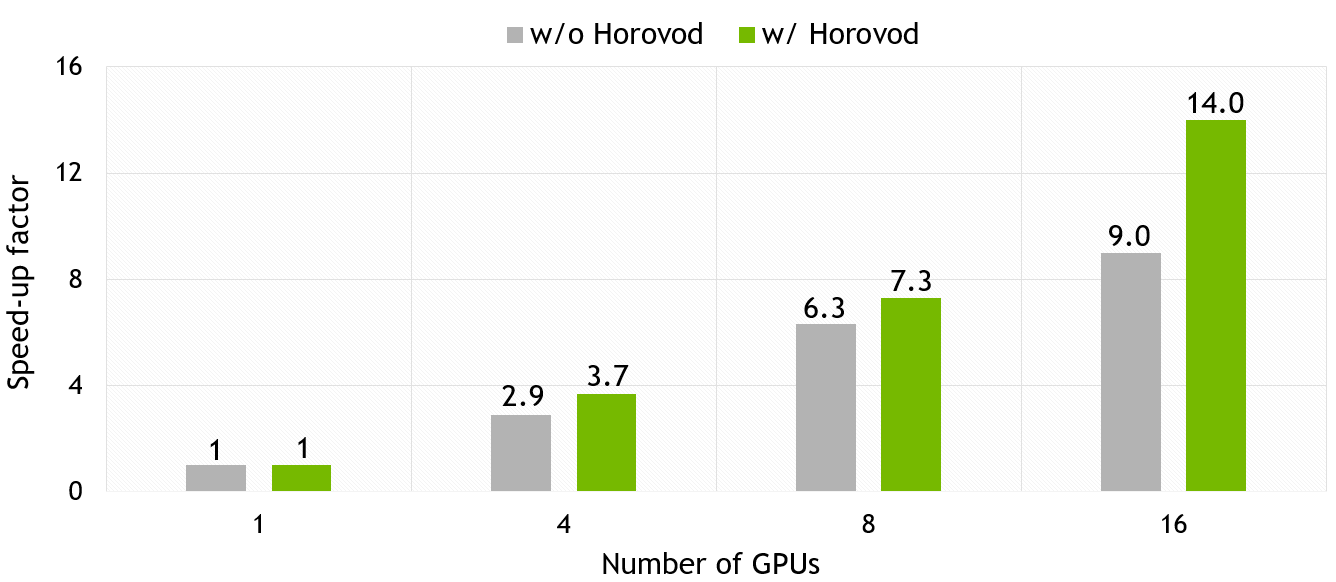}
\caption{Multi-GPU speed-up for ConvS2S}
\label{fig:ConvS2SSpeedup}
\end{figure*}

To use Horovod, you need to install Horovod for GPU,  MPI  and NCCL \footnote{Detailed instructions can be found on Horovod official website at \url{https://github.com/uber/horovod/blob/master/docs/gpus.md}}.  After that, all you need to do is set the parameter \code{use\_horovod} to True in the configuration file and execute \code{run.py} script using \code{mpirun} or \code{mpiexec}. For example:
\begin{lstlisting}[basicstyle=\small]
mpiexec --allow-run-as-root -np <num_gpus> python run.py 
--config_file=... --mode=train_eval --use_horovod=True --enable_logs
\end{lstlisting}
Horovod also allows you to enable multi-node execution. The only thing required from users is to define data “split” solely for evaluation and inference. Otherwise, users write exactly the same code for multi/single GPU or Horovod/Tower cases. 
Horovod gives significantly better scaling for multi-GPU training comparing to TensorFlow native tower-based approach. The specific scaling depends on many factors such as data type, model size, compute amount. For example, the scaling factor for Transformer model is 0.7, while that number for ConvS2S is close to 0.875, as you can see in Figure \ref{fig:ConvS2SSpeedup}.

\section{Models}
\begin{figure*}[ht]
\includegraphics[width=14cm]{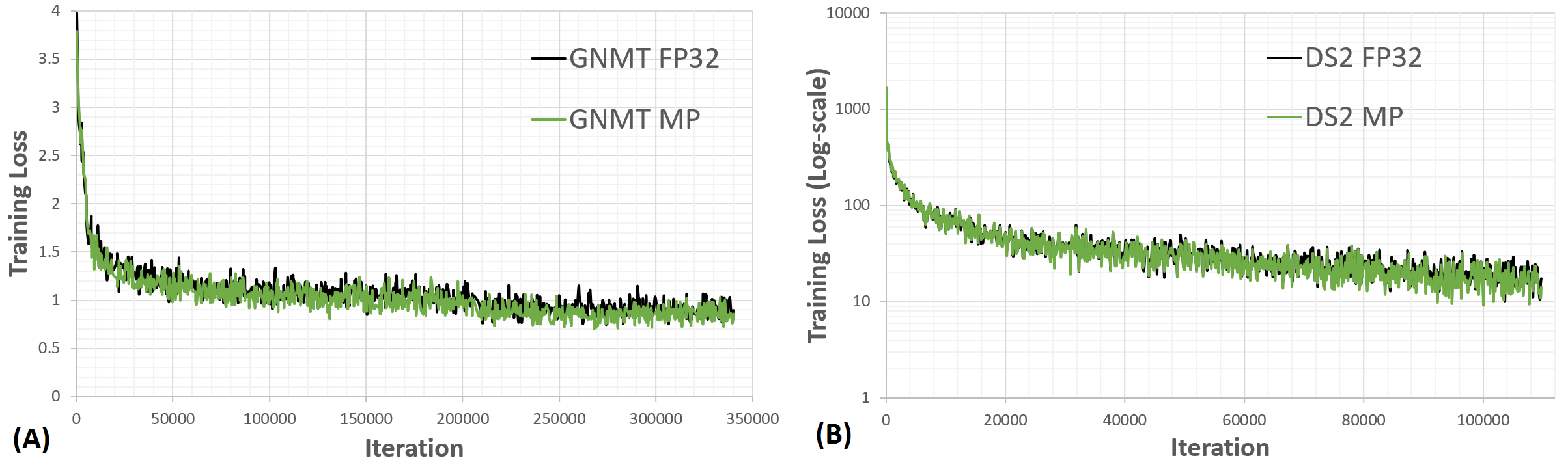}
\caption{\label{fig:losses} Training loss curves for: (A) GNMT-like model, and (B) Deep-Speech-2-like model using FP32 and mixed-precision. For both models, mixed-precision training matches FP32 closely.}
\end{figure*}

OpenSeq2Seq currently offers full implementation of models for the tasks of Neural Machine Translation, Automatic Speech Recognition, Speech Synthesis. On these tasks, mixed-precision training, using the same architecture and hyperparameters as FP32, can speed up training time 1.5-3x without losing model accuracy. Performance boosts vary depending on the batch size. The general rule of thumb is that bigger batch size yields better performance. All configuration files are available on GitHub\footnote{\url{https://github.com/NVIDIA/OpenSeq2Seq}}. Figure \ref{fig:losses} demonstrates that mixed precision has no effect on convergence. 
\subsection{Neural Machine Translation}
Currently OpenSeq2Seq has all the necessary blocks for three models for Neural Machine Translation: (1) Google NMT \citep{wu2016google} (2) Facebook ConvS2S \citep{gehring2017convolutional} (3) Google Transformer \citep{vaswani2017attention}. These blocks can be mixed and matched to create new models.

In our experiments, we used WMT 2016 English$\rightarrow$German dataset obtained by combining the Europarlv7, News Commentary v10, and Common Crawl corpora, resulting in roughly 4.5 million sentence pairs. The scores for these models can be found in Table \ref{tab:wmtBLEU}. These scores are computed using sacreBLEU \citep{post2018call} against \textit{newstest2014.tok.de} file.

\begin{table}[t]
\centering
\begin{tabular}{|c|c|} 
 \hline
 \textbf{Model} & \textbf{sacreBLEU}  \\
 \hline 
 GNMT & 23  \\ 
  \hline 
ConvS2S & 25.0  \\ 
 \hline
Transformer base & 26.6  \\ 
  \hline 
 Transformer big & 27.5  \\ 
 \hline
\end{tabular}
\caption{BLEU scores for different NMT models using mixed-precision training.}
\label{tab:wmtBLEU}
\end{table}

In our experiments, total GPU memory consumption with mixed-precision is reduced to about 55\%, making the training 1.5-2.7x faster comparing to using only FP32. 
\subsection{Automatic Speech Recognition}

OpenSeq2Seq currently has two models for the Automated Speech Recognition task:
\begin{itemize}
\item Wave2Letter+: fully convolutional model based on Wav2Letter \citep{collobert2016wav2letter}
\item Deep Speech 2: recurrent model \citep{Amodei2016DeepS2}
\end{itemize}
These models were trained on LibriSpeech dataset \citep{panayotov2015librispeech} that contains approximately 1000 hours of audio. WERs (word error rates) were measured on \textit{dev-clean} part of the dataset using a greedy decoder - taking at each time step the most probable character without any additional language model re-scoring.
During training in mixed-precision, we observed a total memory reduction to around 57\%, making it 3.6x faster than training using only FP32.
\begin{table}[t]
\centering
\begin{tabular}{|c|c|} 
 \hline
 \textbf{Model} & \textbf{Greedy WER (\%)}  \\
 \hline 
 Wave2Letter+ & 5.4  \\ 
  \hline 
DeepSpeech2 & 6.71  \\ 
 \hline
\end{tabular}
\caption{WERs for ASR models using mixed-precision training.}
\label{tab:wmtBLEU}
\end{table}


\subsection{Speech Synthesis}
OpenSeq2Seq supports Tacotron 2 \citep{shen2018natural} with Griffin-Lim \citep{griffin1984signal} for speech synthesis. The model currently uses only the LJ Speech dataset \citep{ito2017lj}. We plan on additionally supporting the M-AILABS dataset \citep{mailabs}. Sample audio on both datasets can be found on our documentation website\footnote{\url{https://nvidia.github.io/OpenSeq2Seq/html/speech-synthesis/tacotron-2-samples.html}}. Tacotron 2 can be trained 1.6x faster in mixed-precision compared to FP32.
\section{Conclusion and future plans}
OpenSeq2Seq is a TensorFlow-based toolkit that builds upon the strengths of the currently available seq2seq toolkits with additional features that speed up the training of large neural networks up to 3x. It lets users switch to mixed-precision training that takes advantage of the computational power of Tensor Cores available in Volta-based \citep{nvidia2017v100} and Turing-based \citep{NVTuring} GPUs with one single tag. It incorporates Horovod library to reduce training time for multi-GPU and multi-node systems.

OpenSeq2Seq aims to offer a rich library of commonly used encoders and decoders. It currently features a large set of state-of-the art models for speech recognition, machine translation, speech synthesis, language modeling, sentiment analysis, and more to come in the near future as our team is working hard to improve it. Its modular architecture allows quick development of new models out of existing blocks. We plan to extend it with other modules such as text classifications and image-to-text. The entire code base is open-source.

\subsubsection*{Acknowledgments}
We are grateful to Siddharth Bhatnagar and Luyang Chen for their work on previous version of the toolkit. We would like to thank Hao Wu, Ben Barsdell, Nathan Luehr, Jonah Alben, and Ujval Kapasi for their fruitful discussions.

\bibliography{nips_2018}
\bibliographystyle{acl_natbib}
\end{document}